# LAREX – A semi-automatic open-source Tool for Layout Analysis and Region Extraction on Early Printed Books

Christian Reul
University of Würzburg
Am Hubland
D-97074 Würzburg
christian.reul@uni-wuerzburg.de

Uwe Springmann
LMU Munich
Oettingenstraße 67
D-80538 München
springmann@cis.uni-muenchen.de

Frank Puppe
University of Würzburg
Am Hubland
D-97074 Würzburg
frank.puppe@uni-wuerzburg.de

## ABSTRACT
A semi-automatic open-source tool for layout analysis on early printed books is presented. LAREX uses a rule based connected components approach which is very fast, easily comprehensible for the user and allows an intuitive manual correction if necessary. The PageXML format is used to support integration into existing OCR workflows. Evaluations showed that LAREX provides an efficient and flexible way to segment pages of early printed books.

## Categories and Subject Descriptors
Applied computing: document management and text processing – *document analysis, optical character recognition*.

## 1. INTRODUCTION
With the large amount of historical documents that have become available as scanned images in recent years, the run is now open to unlock these heritage data and make them available as fully electronic, machine-actionable research data. Several hurdles must be overcome. The first sub goal is to achieve a high-quality recognition on printed (and manuscript) text with characteristics (unusual glyphs, fonts, background noise etc.) completely different to our modern printings on which existing Optical Character Recognition (OCR) engines have been trained. This has recently been achieved by trained recurrent neural networks with LSTM architecture (see [1] and [2] for printed books with character recognition rates above 95% and [3] for manuscripts). However, a workflow geared at high-volume throughput relies on largely automatic preprocessing methods for the layout analysis task, i.e. the segmentation of pages into text and non-text zones, preferably with semantic meaning (running text, heading, page number, marginalia, etc.). Text zones are further processed and subject to a subsequent OCR step yielding electronic text. If manual work is necessary to obtain a satisfactory segmentation, ease and efficiency is crucial. A third necessary task is the annotation of OCR results in the post correction phase, comprising both error correction and the normalization of the often highly variable historical spelling.

In the following, we present a tool for the difficult page layout analysis task, which is currently the biggest obstacles for a large-scale application of OCR methods to digital images of historical documents. A fast segmentation allowing quick interactive steps to tune global parameters is essential to enable a subsequent automatic turnaround of a whole book, whose single pages can afterwards be corrected individually if needed. Therefore, our semi-automatic approach allows a quick preprocessing compared to purely manual methods and is available[1] under an open-source license to enable the definition of a sustainable, fully open-source OCR workflow.

The remainder of the paper is structured as follows: After a short overview of related work in chapter 2 the functionality of LAREX is described in chapter 3. During the evaluation in chapter 4 the tool is tested in two different scenarios and its performance is measured against Aletheia's. The obtained results are discussed in chapter 5 and chapter 6 concludes the paper.

## 2. RELATED WORK
Given our focus on open-source tools and workflows, it is sensible to divide existing tools into open-source and proprietary software.

Among proprietary tools the *ABBYY*[2] OCR products (Finereader and Recognition Server) define the state of the art of available automatic products with excellent results on a wide variety of layouts. The downside is that they cannot be adapted by the end user. The company does not have any interest in the relatively low volume of historical documents and, less importantly but maybe still of interest, they have a fee assigned. As with any proprietary product long term availability or even adherence to data standards is not guaranteed.

ABBYY is also the underlying engine for automatic preprocessing for the *Transkribus*[3] cross-platform Java tool which is itself open-source, but not the server software behind it.

*Aletheia*[4] (see [4]) is a highly functional proprietary product from PRIMA research group at the University of Salford. It is Windows only and uses Tesseract for automatic preprocessing. Its manual features are very extensive but because of its closed-source nature there is no community contribution except by feature request, no guaranteed long term availability and it comes with a fee.

The open-source OCR engine *Tesseract*[5] has some built-in analysis routines, but is significantly less successful than ABBYY. It is used by Aletheia and also by our own LAREX tool for line segmentation (see section 3.4). The OCR engine *OCRopus*[6] (also open-source) is very capable at segmenting lines of text regions, but offers almost no capability of text/non-text region detection.

The *SCRIBO* module of the OLENA platform (see [5]) is an open-source layout analysis framework which finished second at the 2011 competition on historical book recognition (see [6]). While it clearly has its strengths because of its modularity and flexibility, it does not seem to be suited for the needs of the average user who is looking for a stand-alone tool which can be used right away.

---

[1] https://github.com/chreul/LAREX

[2] http://www.frakturschrift.com/en:start

[3] https://transkribus.eu/Transkribus/

[4] http://www.primaresearch.org/tools/Aletheia

[5] https://github.com/tesseract-ocr/tesseract

[6] https://github.com/tmbdev/ocropy

*Agora* is an open-source software developed in the PARADIIT project[7] of the University of Tours (see [7]). Just like LAREX' it is based on a connected components approach. The software is Windows only and the project appears to be currently dormant.

Finally, Gupta et al. [8] proposed a method that utilizes Tesseract's automatic segmentation in order to perform a text/noise classification on the obtained word bounding boxes. This information is then used to estimate a document's overall quality and increase the OCR accuracy. Despite not performing a layout analysis in the proper sense, this area is still highly relevant for the mass digitization of historical documents. As it is part of eMOP[8] the produced tool has to be free or open-source.

We conclude that an open-source tool that is well documented, adaptable und easily usable is very much lacking and hampering the progress of the OCR of historical books.

## 3. LAREX

The goal of the LAREX (Layout Analysis and Region Extraction) tool currently under development at the University of Würzburg is to support end users in the segmentation and classification of regions in images. In this paper, we discuss its application for analysis of early printed books.

The segmentation procedure is user-driven and uses a connected components approach. It is based on two main assumptions: First, it starts from the premise that related characters, words and text lines usually are closer to each other than unrelated ones. Second, it expects most pages within the same book to have similar layouts or at least to belong to one of only a few layout categories.

LAREX is not designed to be able to automatically segment any given document to perfection. In contrast, it aims to provide the user with a quick and easily comprehensible way to adapt to a given layout, get a segmentation suggestion and to manually correct it if necessary. Therefore, the correction operations should be as simple as possible and deliver an immediate feedback.

### 3.1 Overview

The goal of the proposed approach is to detect and classify different regions on a scanned page (see Figure 1).

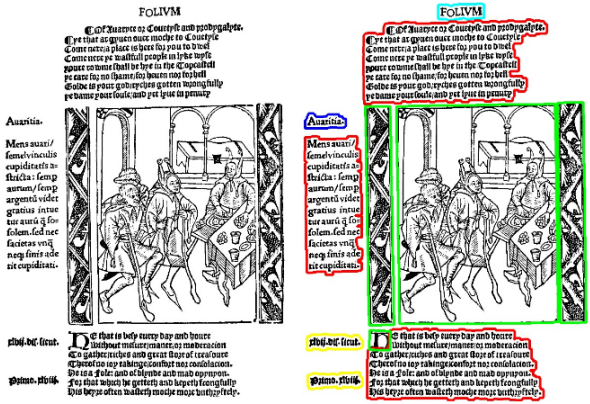

**Figure 1. Example input (left) and output (right) with eleven regions marked belonging to five different types.**

---

[7] https://sites.google.com/site/paradiitproject/home

These regions are connected areas on a page like image, running text (paragraph), headings, marginalia, page numbers etc.. LAREX uses the type definitions of the PageXML format[8] for denoting regions.

The workflow for segmentation consists of the following main steps, illustrated in Figure 2:

Input: Scanned page.
Output: Classified segments as PageXML.
1. Pre Processing: input → resized binary ([A in Figure 2]).
    1.1. Conversion to binary.
    1.2. Definition of a region of interest (optional).
    1.3. Resizing the image.
2. Image detection by Region Growing ([A] → [B]).
    2.1. Region Growing: regions → larger regions (optional).
    2.2. Region Classification: image ↔ no image.
3. Coarse text region detection.
    3.1. Region Growing: regions → larger regions ([C]).
    3.2. Region Classification using constraints on attributes of regions: regions → region types ([C] → [D][E]).
    3.3. Post-processing: pruning and fine tuning of region types.
4. Manual amendments and text sub region classification (optional, see sections 3.3 and 3.4) ([E] → [F]).
5. Conversion to output format.

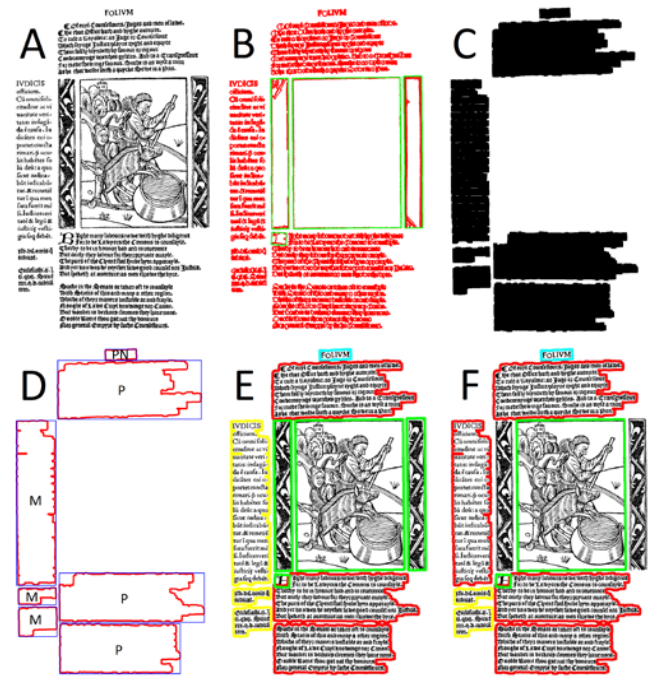

**Figure 2. Workflow. A: binary, B: after image detection, C: after image removal and region growing, D: after text region classification (P: paragraph, M: marginalia, PN: page number), E: algorithm output, F: after user amendment.**

### 3.2 Detailed Description of the Algorithm

In this section the steps of the basic workflow are described. Optional manual adaptations and the more advanced text sub-region detection are discussed separately in section 3.3 and 3.4.

---

[8] http://www.primaresearch.org/tools/PAGELibraries

### 3.2.1 Preprocessing

LAREX accepts color, grayscale and binary images as input. However, the upcoming region growing and contour detection operations require the binary format. Therefore, a conversion is performed if necessary. Obviously, the resulting segmentation still corresponds to the original input image. The user can work on the entire binary image or specify a region of interest by drawing a rectangle, e.g. to exclude artefacts of the input image due to scanning. Everything outside the specified area is painted white and will not have any effect on the rest of the segmentation process. A region of interest can either be specified for a single page or be applied for the entire book. It can be adjusted at any time. The image is resized to a given resolution (default: 1600 pixels in height) in order to speed up the segmentation process and to normalize parameters like the upcoming dilation catchment areas or the required minimum area of the regions.

### 3.2.2 Image Detection

Images are expected to have a certain minimum area and to either be very compact or to possess a border. First, a dilation operation is applied using small kernel values to prevent letters and words from merging as well. It may be imagined as a growing process during which black foreground pixels within a defined kernel grow together. All connected components bigger than the given image area threshold (default: 3000 pixels) are marked as images. For further processing either only the contours themselves or the straight or rotated bounding rectangles around them are removed from the image.

### 3.2.3 Coarse text region detection

Afterwards, another dilation process with a bigger kernel than before takes place (step 3.1). Then in step 3.2, the newly emerged regions are assigned a type by enforcing several constraints. The constraints for type classification (Table 4) use the region attributes shown in Table 1, constants in Table 2 and functions in Table 3.

**Table 1. Region attributes.**

| Attribute | Description |
|---|---|
| contour | A list of points describing the outer contour of the region. |
| rectangle | The straight bounding rectangle of the contour. |
| area | Area of the contour (in pixel). |
| max-occurrence | Maximum valid number of occurrences of a type on a page: {1, unbounded} |
| priority-position | Preferred positions if max occurrence = 1: {top, bottom, left, right} |

**Table 2. Constants.**

| Constant | Default |
|---|---|
| minAreaParagraph | 2000 px |
| minAreaMarginalia | 2000 px |
| minAreaPageNumber | 500 px |

**Table 3. Functions.**

| Function | Description |
|---|---|
| >, <, =, … | Returns true / false |
| within(rectangle, rectangles) | Returns true if rectangle is entirely covered by one of the rectangles. |

**Table 4. Constraints of the default text types.**

| Type | Constraints |
|---|---|
| paragraph | region.area > minAreaParagraph |
| marginalia | region.area > minAreaMarginalia, within(region.rectangle, marginalia.rectangles)[9] |
| page number | region.area > minAreaPageNumber, within(region.rectangle, pageNumber.rectangles)[10], max-occurrence = 1, priority-position = top |

Table 4 shows constraints of three region types: paragraph, marginalia and page number. Position constraints of different regions may overlap. For all types it is checked if the area of the region is bigger than the type-specific minimal area. If this is the case and if the bounding rectangle of the region is completely covered by the rectangle-position of the type, the region is considered a candidate for that type. During the initial classification step regions can be assigned several types. In the post processing step (3.3.) a type priority list is applied. The default order is first page number, second marginalia, and third paragraph. Each region that satisfied the constraints of two or more types is assigned to the type with the highest priority. In addition, it is possible to define the number of maximal occurrences on a page for each type. For example, usually there exists only one page number. Therefore, only the best fitting page number candidate is chosen and the others are either re-assigned to other types or dropped. Again, for this decision a priority list is used. In case of the page number the priority position is 'top', therefore, the topmost candidate is chosen.

### 3.2.4 Conversion to output format

When the user is satisfied, the segmentation result can be stored as an XML file using the PageXML format. Thus, the segmentation result is recalculated to the true size of the input image. Afterwards the resulting file is placed within the same folder as the input using the same file name.

## 3.3 Manual Adaptions

In many cases the default parameters will not be enough to obtain a satisfying segmentation result. In the following sections global and local operations to improve the results are shown.

### 3.3.1 Global Adjustments to the given Book

Figure 3 shows an example outcome of the default setup consisting of four regions (*image*: green, can occur anywhere on the page; *paragraph*: red, anywhere; *marginalia*: yellow, left and right 25%; *page number*: cyan, top and bottom 25%). Obviously, there is still some work to be done. The initial was correctly detected as well as the page number. However, the marginalia regions got classified as paragraph, as their outer contour was not located entirely within the marginalia regions. Furthermore, there is a signature mark at the very bottom of the page. Lastly, the vertical spacing between text lines is relatively big, leading to the unwanted separation of text blocks. It is worth mentioning that the marginalia located at the bottom right is not classified as a page number, despite being in an eligible position. That is because by default the max occurrence value of the page number region is 1 and the priority position is top.

The default setup can be easily adapted by only a few clicks. First, the page number position at the bottom is deleted and the upper one is slightly adjusted, as in this book page numbers always occur at the top center of the page. Furthermore, the marginalia regions have to be expanded towards the middle of the image. The dilation parameters are slightly increased. Finally, another region of the type *signature-mark* (maroon) is added and located at the bottom center of the page. The settings can be saved and applied to the rest of the book as well as to other books with similar layouts.

---

[9] Default case for marginalia: rectangle has to be located within the left or right 25% of the page.

[10] Default case for page number: rectangle has to be located within the top or bottom 25% of the page.

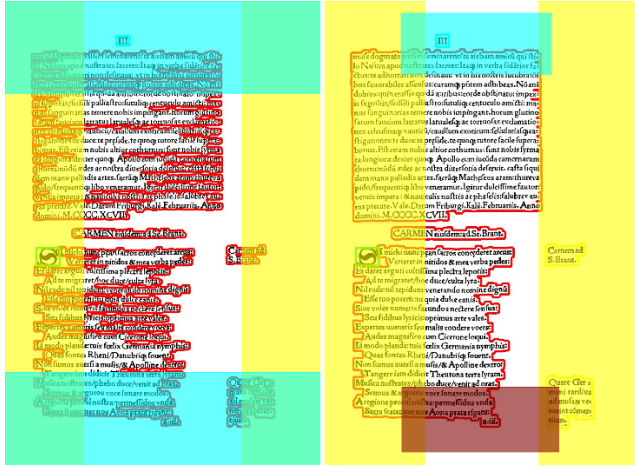

**Figure 3. Result using the default (left) and slightly adjusted (right) parameters.**

### 3.3.2 Local Manual Corrections

For special cases the user can make amendments like adjustment of region parameters or the manual adaption of the obtained segmentation result by deleting segments, changing their type and splitting or merging them. For example, in Figure 3 the signature mark is connected to the bottom text block which is difficult to correct with global parameter optimization. There are several ways to deal with such problems (see Figure 4): The segments can be cut by drawing a line or a string of lines. The separated block gets automatically assigned the correct type if it is within the corresponding position. As an alternative, segments can be marked by drawing a rectangle or a polygon around them and assigning a type. This remains fixed for the entire segmentation process and is unaffected by other parameters.

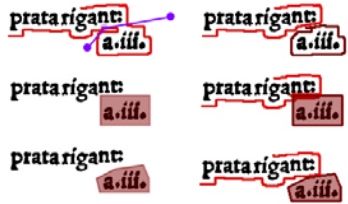

**Figure 4. Three ways to influence the result directly (left: user action, right: resulting segmentation): by drawing lines (top), a rectangle (middle) or a polygon (bottom).**

## 3.4 Text Sub-Region Segmentation

Sometimes it is necessary to cut a detected text block along one or several of its lines, e. g. when a heading or a highlighted first line of a paragraph has to be cut from the rest of the text. This can be done by drawing a cut line. If a lot of line segmentation is needed, one can utilize Tesseracts page segmentation algorithm and manually correct rather than do it completely manually.

### 3.4.1 Locating the Lines

To assure optimal line segmentation results the previously detected segments are processed separately. After cutting them from the original image, they are rotated into an upright position. If a segment is too small for Tesseract to predict the needed angle, the weighted average of the remaining segments is used. Now the line segmentation can take place. Afterwards, the resulting lines are rotated back into their original position.

This process can be quite time consuming: During a test on an image of the size 5100 x 7800 pixels it took a PC (i7-6700) with 3.40 GHz around 20 seconds to complete the line segmentation. Reducing the image resolution usually worsens the result significantly. Therefore, it is recommended to collect an arbitrary number of temporary segmentation results and then process all of them at once, for example during night time. The resulting line coordinates are stored in a separate PageXML file and automatically matched to the original segmentation result at runtime, as soon as the corresponding page is selected in the tool.

An example output of the line segmentation can be seen in Figure 5. Despite the heavy skewing of the image, the result is very precise. When the mouse cursor is moved over one of the block segments, the corresponding lines are displayed and the one directly under the cursor is marked as active.

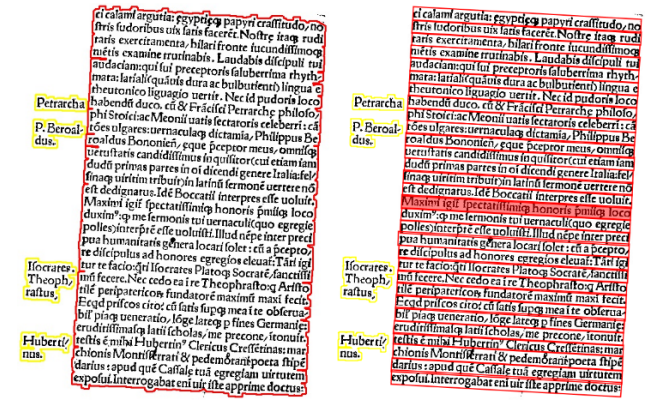

**Figure 5. Example block (left) and line segmentation (right).**

### 3.4.2 Putting the Line Segmentation to Use

While the manual separation of a text block is a tedious task, it is much easier when using the text lines. After selecting a line with a mouse click the user can cut the underlying segment above or below the selected line. A third option is to change the type of the line. This automatically triggers both cuts.

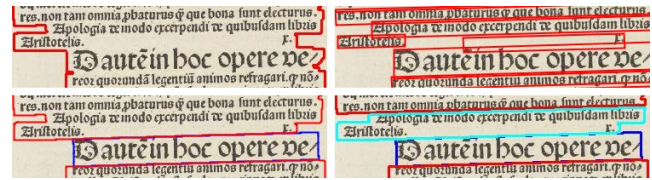

**Figure 6. Example operational scenario for applying the line segmentation: segment (top left), detected lines (top right), extracted first line (bottom left), final result (bottom right).**

Figure 6 shows the line segmentation in action. In this particular book each chapter starts with a heading which looks like standard text but is slightly indented. It is followed by the first line of the paragraph which is printed in a bigger, more prominent font. The standard segmentation process recognizes a single block (top left), as the gaps around these special parts of the text are equal to the ones of the standard text. The task is to split these segments from the text and assign the correct types to them. This can be done by a few simple actions: After the line segmentation (top right) the first line of the paragraph is marked as the desired type (bottom left; blue) triggering the line cutting. Next, the heading is split from the text above and the resulting text block is marked as heading (bottom right; cyan).

## 4. EVALUATION

Due to its interactive design it would be misleading to compare LAREX to approaches by using evaluation metrics for fully automatic tools. In addition, for an interactive tool the overall required time and user effort to achieve a satisfactory segmentation is important. Therefore, both the number of problems and the time needed to correct them were assessed. We report two evaluations: The first one with the standard tool and the second one with slight extensions for special requirements. Additional to the time exposure the OCR accuracy on the resulting text regions was compared to the one obtained by fully manual segmentation.

### 4.1 Evaluating the Standard Functionality

The first evaluation was performed on 'The Shyp of Folys' (The Ship of Fools)[11] as part of an effort to support the 'Narragonien digital' project[12]. The book consists of 570 pages and alternates between English and Latin text. Figure 7 shows the segmentation guidelines for this task: Images and initials have to be marked, while the ornaments around the image are supposed to be dropped. All text regions, marginalia, signature marks and page numbers/sheet titles have to be recorded. The image descriptions are supposed to be classified as normal text.

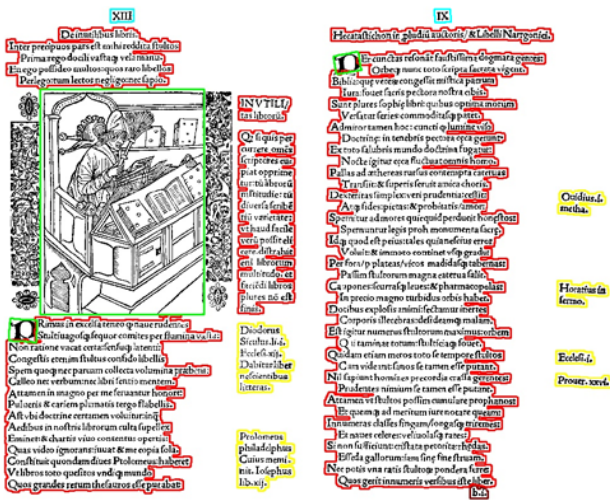

**Figure 7. Expected segmentation examples.**

The segmentation process was performed by a student research assistant of the Narragonien digital project who had some prior experience with LAREX but no background in computer science, image processing or layout analysis. In total, 378 corrections were performed on the first 200 of the 570 scanned pages (see Table 5).

The vast majority of user actions consisted of the removal of unneeded regions like ornaments, stains or the bordures next to the main images. Not all corrections were due to real segmentation errors. A closer analysis showed that about half of the amendments became necessary due to additional requirements, initially not specified in the parameter setup. This included the removal of bordures next to images or of ornaments at the end of text blocks or lines. Furthermore, LAREX classified text blocks next to images (e.g. Figure 7, left) as marginalia whereas they are supposed to be running text. Instead of trying to optimize the parameter setup in order to cover all occurring special cases it is often more efficient to simply correct the respective annotations manually.

**Table 5.** Required user actions on 200 pages.

| Description | # | Effort |
|---|---|---|
| Removing irrelevant regions | 161 | Right click |
| Removing irrelevant images | 90 | Right click |
| Type correction: marginalia → image description | 65 | Double click and type |
| Cutting off signature-mark | 31 | Draw lines |
| Type correction: paragraph → marginalia | 17 | Double click and type |
| Combining text blocks | 10 | Rectangle and type |
| Manual labeling of images | 2 | Rectangle and type |
| Manual labeling of the page number | 2 | Rectangle and type |

Fortunately, the most frequent errors requiring individual adjustments are most easily to correct. Almost two thirds of the applied corrections required only a single mouse click on top of the region to resolve the issue. Furthermore, close to 90% of the alternations can be done by either a single click or by a double click followed by a selection of the desired region type. The costlier corrections are rare requiring fitting lines, rectangles or polygons into the text.

For the layout analysis of the entire book a total time expenditure of 2 hours and 18 minutes was recorded, including the time required for finding an optimal parameter set. Within the first 30 minutes 80 scans were segmented. Subsequently, the number of processed pages per half hour rose to 112, 131 and 149. The last 98 scans were processed within 18 minutes (≙ 163 scans per 30 minutes).

For comparison, a student research assistant of the University Library of Würzburg managed to segment only 160 pages (28% of the entire book) during the same 2 hours and 18 minutes using Aletheia. It is worth mentioning that the student had extensive experience in segmenting early printed books with Aletheia.

### 4.2 Application to 'Der Heiligen Leben'

LAREX was also deployed to the book 'Der Heiligen Leben', printed by Anton Koberger in Nuremberg in 1488[13]. For an in-detail description see Reul et al.: Case Study of a highly automated Layout Analysis and OCR of an incunabulum: 'Der Heiligen Leben' (1488) (submitted to this conference). Figure 8 shows a representative selection of page layouts occurring in the book.

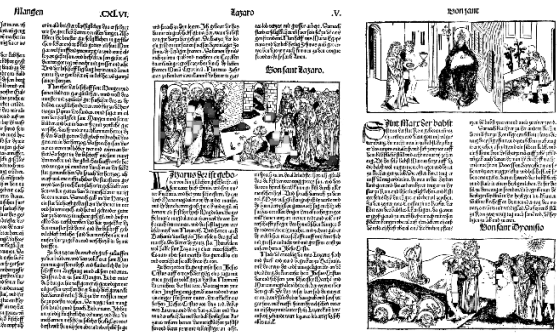

**Figure 8. Example pages from 'Der Heiligen Leben'.**

A slight alteration of the default setup was necessary to achieve an appealing coarse segmentation. However, at this point several headings, i.e. lines consisting of more prominent letters, were often still connected to or even surrounded by running text. To detect and separate these segments, the line location and segmentation functionality described in 3.4 was utilized. Additionally, simple rules were applied taking the height of a heading line candidate and the area of its connected components into consideration. Moreover, a reading order was determined for each page. Finally, the segmentation results were manually checked and corrected if necessary.

---

[11] Translated by Alexander Barclay, 1509.
   Scans provided by the University Library of Würzburg.

[12] http://kallimachos.de/kallimachos/index.php/Narragonien

[13] http://vb.uni-wuerzburg.de/ub/itf954/ueber.html

During the evaluation, it was shown that the OCR accuracy decreased slightly from 97.57% (Aletheia) to 97.35% (LAREX). However, the overall time expenditure needed for segmentation was reduced significantly by using LAREX. While the (semi-)automatic extraction finished in less than six hours (including manual tuning for global properties), the fully manual extraction lasted close to 100 hours. Notably, an additional correction of the LAREX output within the tool took almost two hours but improved the OCR accuracy only insignificantly (97.37%). This implies that the raw text/non-text segmentation worked very well.

## 5. DISCUSSION

The evaluation illustrated that LAREX is a powerful tool for the segmentation of early printed books due to its interactivity and flexibility. The assessment of the required user actions illustrated that LAREX enables the user to influence the result heavily with minimal effort and without changing the setup in itself. For example, a common error is the non-separation of small text blocks like the signature mark or the catchword from the running text. These types of errors would require considerable effort to be prevented by adjusting the expected region positions or the dilation kernel, as there is no real spatial separation from the adjacent region. However, it is very easy to correct these mistakes manually. Assessing and optimizing such trade-offs is part of our future work.

The application on 'Der Heiligen Leben' showed that further functionality can dramatically reduce the required user effort while only slightly decreasing the OCR accuracy. During the experiment additional rules had to be added within the program code. For a more generic approach a larger set of region or line attributes as well as rules utilizing them are required. Moreover, the user has to be able to enter rules directly, test a setup and change it if necessary.

## 6. CONCLUSION AND FUTURE WORK

A semi-automatic tool for layout analysis and region extraction was proposed. LAREX uses parameters in order to be flexible enough to be adjusted to different book layouts. It offers maximal value on books with an extensive number of pages whose layouts are complex but consist of clearly spatially separated segments. The manual corrections can be done in a simple and time efficient way. LAREX does not offer an almost pixel-perfect segmentation like Aletheia. However, as the experiments on 'Der Heiligen Leben' showed, there are cases where a much faster but slightly less accurate segmentation may be completely sufficient.

Despite LAREX already being a fully functional tool, there are still many things that could be added or refined to further improve it: First, the rule functionality will be extended as described in chapter 5 to offer the user more parameters and to make the tool more flexible. Furthermore, the post-processing and correction performance could be improved by providing the user with further tools to perform an even finer grained segmentation if needed. In addition, many smaller additions are planned, e. g. user-driven swash capital removal and optimized line segmentation. The latter could further improve the text sub-region detection or even directly support OCR. However, the main goal remains the improvement of the declarative specification of rules based on a range of extracted parameters by image processing operations. To enhance LAREX´ range of functions considerably while keeping most of its simplicity will be a challenging task for the future. Due to its open-source nature, LAREX will benefit a lot from the anticipated feedback and co-development of a global user group. The ready-to-use tool together with some example data are available at https://go.uniwue.de/larex.

We want to highlight that (to our knowledge) LAREX is currently the only available fully stand-alone open-source tool under active development that significantly reduces the time of manual high-quality page analysis for historical books. It has been shown to be usable by non-specialists after a very short training period. Hopefully, it will help to break down the preprocessing barrier for mass-OCR of digital images.

## 7. ACKNOWLEDGMENTS

The authors would like to thank Maximilian Wehner and Maximilian Nöth for their testing efforts, Marco Dittrich and Martin Gruner for their helpful remarks on the tool and the University Library of Würzburg for providing the required scans.